\documentclass[11pt]{article}

\usepackage[final]{acl}
\usepackage{times}
\usepackage{latexsym}
\usepackage[T1]{fontenc}
\usepackage[utf8]{inputenc}
\usepackage{microtype}
\usepackage{inconsolata}
\usepackage{graphicx}
\usepackage{amsmath}
\usepackage{array}
\usepackage{booktabs}
\usepackage{xcolor}

\title{Why Pay for Long Reasoning When Rewrite Do Trick?\\[2pt]
\large Causal Effects of Reasoning History Compaction on Future Generation and Agent Cost}
\author{{\fontsize{14}{16}\selectfont\rmfamily\bfseries Anton Hristov} \\
{\fontsize{14}{16}\selectfont\rmfamily\mdseries Technological School Electronic Systems} \\
{\fontsize{14}{16}\selectfont\rmfamily\mdseries hristov.anton@icloud.com}}
\hypersetup{pdfauthor={Anton Hristov},pdftitle={Why Pay for Long Reasoning When Rewrite Do Trick?}}

\begin{document}
\maketitle

\begin{abstract}
Reasoning substantially improves language models' ability to solve difficult
problems, but long reasoning traces make these gains expensive. Efficient reasoning should express useful analysis in fewer
tokens by reducing repetition and text that adds little value to solving the
problem. Progress in proprietary models motivates this
goal, yet their hidden scratchpads limit direct study of how it is achieved.
We investigate whether open-weight models can generate less verbose reasoning
by changing the form of their own earlier thoughts, without retraining.
Our intervention rewrites prior reasoning into compact state records before
pending tool results are revealed, then compares continuations from original
and compact histories. Across the tested models, accepted compact histories
usually shorten future reasoning; a coding case study also shows less repeated
verification after a shared candidate solution. In software-agent deployments,
compaction can reduce total spending, although task quality and rewrite-inclusive
time depend on the target and compressor. These findings identify reasoning
history as one way to reduce future reasoning-token use and motivate
model-specific compaction policies evaluated for both efficiency and retained
capability.
\end{abstract}

\section{Introduction}
\label{sec:introduction}
Reasoning is a major source of language-model capability: generating intermediate
steps improves problem solving, and recent reasoning models extend these gains
to complex coding and tool use \citep{wei2022chain,openai2026gpt55}.
Those gains carry a substantial inference cost.
In Artificial Analysis data retrieved on August 3, 2026,
reasoning accounts for 84.9\% of generated tokens for
GLM 5.2 and 75.1\% for Claude Fable 5. For DeepSeek V4 Pro, reasoning output
alone accounts for 53.4\% of total API cost \citep{artificialanalysis2026}.
Reasoning can therefore dominate generated output, while agents also repeatedly
pay to process earlier scratchpads as input history.

The goal of efficient reasoning is to retain the analysis needed to solve a
problem while reducing repeated analysis and low-value tokens.
OpenAI reports that GPT-5.5 improves coding performance over GPT-5.4 while using
fewer tokens \citep{openai2026gpt55}. Practical inspiration for this study came
from occasional, unofficial exposures of GPT-5.5's reasoning scratchpad, which
appeared compact (Appendix~\ref{app:motivating-observations}). These observations are anecdotal, and the scratchpads are not
reliably accessible for controlled study: OpenAI does not officially expose raw
reasoning \citep{openai2026reasoning}. We therefore study open-weight models whose
reasoning is accessible for observation and controlled rewriting.

Open-weight models may already be capable of expressing useful
reasoning compactly, even when their default responses are verbose.
We test whether rewriting earlier scratchpads as concise state
records can encourage these models to use less reasoning without new training. Prior reasoning may serve as an
in-context example of how concisely to express subsequent analysis. Changing its form could
therefore reduce newly generated reasoning, beyond the immediate saving from
shortening the input. The rewrite is instructed to preserve decisions,
uncertainty, and intended actions; the target receives no new instruction to be
brief.

Concise demonstrations and surrounding context already influence reasoning
length \citep{jin2024impact,renze2024benefits,xu2025chain,ge2025innate}.
Reasoning Shift shows that added context can shorten reasoning and verification,
sometimes at a quality cost \citep{rodionov2026reasoning}. Related approaches
manage agent context \citep{sun2025contextfolding,lian2026sweagile,li2026selfcompacting}
or steer concise reasoning states \citep{azizi2025activation,chen2025seal}.
We isolate a rewrite of the model's own earlier reasoning while preserving
other messages and withholding future tool results from the compressor.

We first test whether this intervention changes subsequent generation in paired
continuations, then evaluate continuous compaction in software agents.
The controlled results show frequent shortening across model families, while
deployment quality and economics vary by target and compressor. Compaction is
thus a candidate route toward more efficient reasoning, with practical value
that must be assessed per model rather than inferred from shorter traces alone.

\begin{figure*}[t]
\centering
\includegraphics[width=\textwidth]{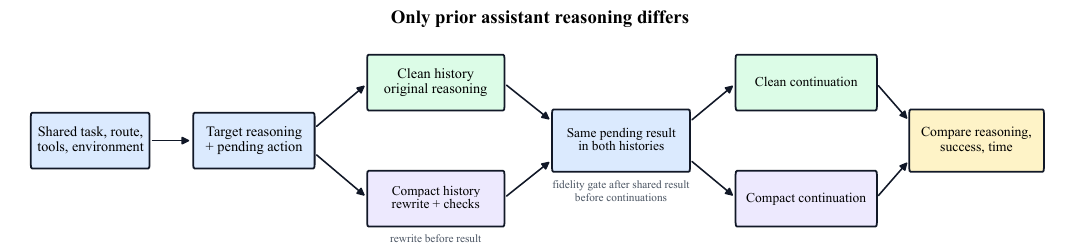}
\caption{Outcome-blind paired fork. The target requests an action, its reasoning
is rewritten before the result is available, and both histories receive the same
result. Fidelity acceptance is required before continuation sampling. Only prior
reasoning differs at the first continuation; later actions and observations can
diverge.}
\label{fig:protocol}
\end{figure*}

\section{Method}
\label{sec:method}
\paragraph{Reasoning-only compaction.}
A compressor LLM converts historical assistant scratchpads into shorter state
records. It is instructed to retain decision-relevant facts, numbers,
uncertainty, conclusions, and intended actions while removing repetition.
System and user messages, tool calls, arguments, and observations remain intact.
The target's weights, tools, and current instructions are unchanged.
We compare original (\emph{clean}) and rewritten
(\emph{compact}) histories; Table~\ref{tab:compaction-example} gives an example.
Code and frozen experimental records for reproducing the analyses are
released at \url{https://github.com/DebelToni/ShortReasoning}.

\paragraph{Outcome-blind timing.}
During shared acquisition, each rewrite completes after an action is requested
and before its result is executed or revealed. At the fork, the histories are
frozen and independently continued under the same model/provider route and
settings (Figure~\ref{fig:protocol}). Serialized first requests are checked
field by field to verify that only assistant reasoning differs. Consequently,
the first-turn contrast measures the effect of the history intervention after a
shared observation; later contrasts can also reflect treatment-induced
trajectory changes.

\paragraph{Acceptance and estimand.}
Routes must demonstrate functional use of replayed reasoning. Action and fidelity
gates select sources before continuation. GPT 5.6 Sol writes rewrites and judges
them in separate rubric-based calls. The estimate is conditional on accepted
treatment histories (Appendix~\ref{app:protocol}).

We measure provider-exposed future reasoning tokens.
For clean count $R_C$ and compact count $R_K$, we report $100(R_K/R_C-1)$;
negative values indicate reduction. Token comparisons use completed pairs with
equal postfork turn counts. This endpoint-dependent matching can select on a
treatment consequence. Exact success retains every included pair. Confidence
intervals resample source histories with their continuation replicates.

\section{Experiments}
\label{sec:experiments}
\subsection{Controlled Continuations}
\label{sec:controlled}
\paragraph{Effect on subsequent reasoning.}
We test whether compacting earlier reasoning changes how much reasoning the
model generates later on. Six model/provider routes
each attempt 12 multi-stage tasks with deterministic tool responses and exact
final-answer checks. For example, a rescue-planning task requires choosing a
plan under cost, site-count, reliability, and risk constraints as tools reveal
simulations and a superseding weather update. A source history is one model's
partially completed run on one task.
Precontinuation gates retain 58 of 72 model-specific source histories: five
fail action checks and nine fail fidelity checks. Three paired continuations
per accepted source yield 174 quality pairs and 172 comparable token pairs
(Table~\ref{tab:controlled}). Three length-capped clean continuations and one
compact provider-error continuation were rerun; Appendix~\ref{app:first-attempt}
reports the retry protocol and first-attempt sensitivity.

\begin{table}[t]
\centering
\scriptsize
\setlength{\tabcolsep}{1.5pt}
\resizebox{\columnwidth}{!}{%
\begin{tabular}{@{}lrrrr@{}}
\toprule
Route & \shortstack{Sources\\retained} & \shortstack{Compact\\shorter} & Median $\Delta$ [95\% CI] & \shortstack{Success\\clean$\to$compact} \\
\midrule
DeepSeek V4 Flash & 12/12 & 30/36 & $-29.5\%$ $[-39.5,-16.5]$ & 36/36$\to$35/36 \\
Laguna S 2.1 & 12/12 & 32/36 & $-46.2\%$ $[-56.3,-29.6]$ & 36/36$\to$35/36 \\
MiniMax M3 & 10/12 & 25/30 & $-27.3\%$ $[-38.4,-17.1]$ & 30/30$\to$30/30 \\
GLM 4.7 Flash & 5/12 & 13/13 & $-50.5\%$ $[-66.6,-40.1]$ & 11/15$\to$14/15 \\
GLM 5.1 & 10/12 & 28/30 & $-20.2\%$ $[-28.6,-12.5]$ & 25/30$\to$25/30 \\
DeepSeek V4 Pro & 9/12 & 23/27 & $-32.6\%$ $[-43.4,-12.2]$ & 27/27$\to$27/27 \\
\bottomrule
\end{tabular}%
}
\caption{Retrospective retry-augmented results on accepted histories. Shorter
counts use equal-horizon pairs; success includes all three replicates per source.
Intervals use 10,000 source-level bootstrap resamples. Retention differs across
routes, precluding a matched-task model ranking.}
\label{tab:controlled}
\end{table}

\begin{figure}[t]
\centering
\includegraphics[width=\columnwidth]{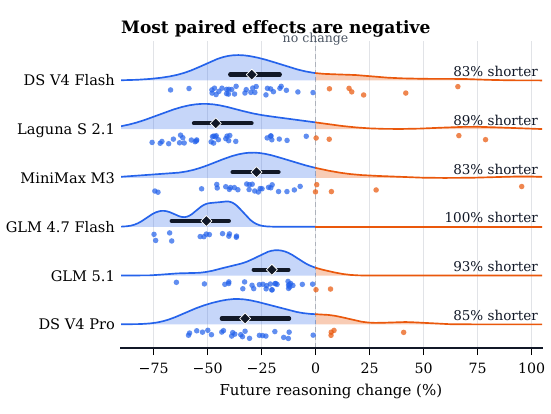}
\caption{Paired future-reasoning changes across the six routes (172 comparable
pairs). Points show continuations; diamonds and bars show medians and
source-clustered 95\% intervals. Blue marks shortening; orange marks increases
or equality. DS denotes DeepSeek.}
\label{fig:raincloud}
\end{figure}

Compaction shortens 151/172 comparisons (87.8\%), with route medians
between $-20.2\%$ and $-50.5\%$. Twenty pairs grow longer and one is
unchanged (Figure~\ref{fig:raincloud}); the median absolute change is
$-178$ tokens, with a maximum increase of 2,807. Exact success changes from
165/174 to 166/174: GLM 4.7 Flash gains three successes, DeepSeek V4 Flash and
Laguna each lose one, and other routes have unchanged totals. Only the most
heavily filtered route gains. The evidence supports frequent
shortening within accepted histories, with route-dependent quality changes.

\begin{figure}[t]
\centering
\includegraphics[width=\columnwidth]{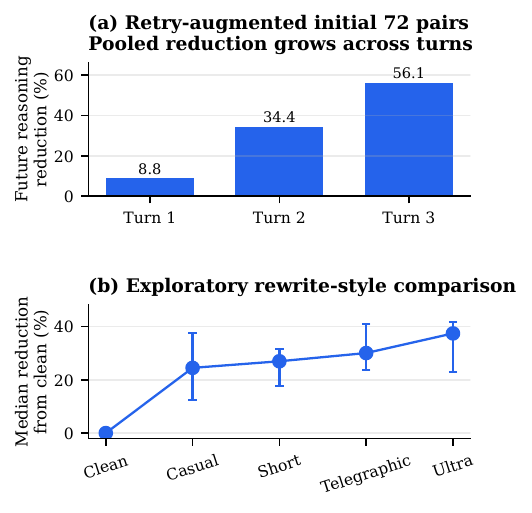}
\caption{Two views of later reasoning. (a) Pooled reductions
on the retry-augmented initial 72 pairs. (b) Median reductions and source-level
95\% bootstrap intervals across exploratory history styles, with 30
continuations per arm. The panels use different source sets and summaries.}
\label{fig:controlled-summary}
\end{figure}

\paragraph{Turnwise and style patterns.}
Figure~\ref{fig:controlled-summary}(a) examines shortening across
turns in the initial DeepSeek/Laguna comparison (72 pairs). Pooled token
reductions grow from 8.8\% on turn one to 56.1\% on turn three; later turns
can also reflect changes in previously generated text and actions. Panel (b)
compares five history styles on ten DeepSeek sources, with three continuation
blocks each. More compact styles have larger median reductions, although
individual responses are not consistently ordered. This sweep uses sources
with known outcomes and jointly varies length, syntax, wording, and compressor
provenance, leaving the mechanism unresolved (Appendix~\ref{app:dose}).

\paragraph{Human fidelity assessment.}
Two nonexpert raters independently assessed six shared
original/rewrite pairs, one per route. Six longer third-turn pairs were added
after rater A's first submission and are reported separately
(Table~\ref{tab:human-fidelity}). Disagreements concerned phase wording and the
intended next action; one response also showed uncertainty about a task term.
These findings have limited scope (Appendix~\ref{app:human-fidelity}).

\begin{table}[t]
\centering
\begingroup
\small
\setlength{\tabcolsep}{4pt}
\begin{tabular}{llrrr}
\toprule
Batch & Rater & Pass & Problem & Unsure \\
\midrule
Easy & A & 6 & 0 & 0 \\
Easy & B & 4 & 1 & 1 \\
Longer & A & 6 & 0 & 0 \\
Longer & B & 5 & 1 & 0 \\
\bottomrule
\end{tabular}
\endgroup
\caption{Human fidelity labels. Overall-label agreement is
4/6 for the easy initial batch and 5/6 for the longer,
post-start, complexity-selected extension. Original judgments are retained without adjudication.}
\label{tab:human-fidelity}
\end{table}

\paragraph{Rewrite properties and length-matched controls.}
To test whether generic rewording elicits shorter continuations, we compare
original histories with fresh paraphrases constrained to 90--110\% of their
word count. Across 65 comparable pairs, median future reasoning changes
by $+0.7\%$ (95\% CI $[-8.2,+9.7]$), while exact success falls from 68/69 to
59/69. The near-zero median change gives little evidence of systematic
shortening from this paraphrase intervention. Its lower success rate leaves the
specific contribution of history length unresolved.

\paragraph{Verification behavior in a coding case study.}
To examine checking after a candidate solution is available, we compare clean
and compact continuations with the same recorded program and identical
public-test feedback. We measure reasoning on the submission turn. In the
LiveCodeBench screen \citep{jain2024livecodebench}, extended verification,
defined retrospectively as at least 350 reasoning tokens, occurs in 61.1\% of
successful DeepSeek clean branches and none of the compact branches. This
illustrates shorter deliberation before submission in a selected successful
setting (Appendix~\ref{app:code}).

\begin{table*}[t]
\centering
\scriptsize
\setlength{\tabcolsep}{2.1pt}
\begin{tabular}{@{}llrrrrrrrr@{}}
\toprule
& & & \multicolumn{4}{c}{Unique reasoning-history blocks} & \multicolumn{3}{c}{Change from Normal (\%)} \\
\cmidrule(lr){4-7}\cmidrule(l){8-10}
Target & Compressor & Score & Raw words & Compact words & Saved words & Reduction & Spend & Tool calls & Target time$^{*}$ \\
\midrule
DeepSeek V4 Flash & External & 32/50 & 168,611 & 83,851 & 84,760 & 50.3\% & $-34.5$ & $-19.9$ & $-34.2$ \\
 & Self & 29/50 & 158,476 & 68,423 & 90,053 & 56.8\% & $-46.5$ & $-29.8$ & $-43.5$ \\
MiniMax M3 & External & 31/50 & 146,126 & 55,901 & 90,225 & 61.7\% & $-34.0$ & $-23.6$ & $-62.9$ \\
 & Self & 30/50 & 131,026 & 49,662 & 81,364 & 62.1\% & $-38.2$ & $-24.8$ & $-60.8$ \\
Nemotron 3 Ultra & External & 32/50 & 129,834 & 74,037 & 55,797 & 43.0\% & $+9.4^{\dagger}$ & $+7.7$ & $-27.5$ \\
 & Self & 29/50 & 146,445 & 62,553 & 83,892 & 57.3\% & $+7.2^{\dagger}$ & $+0.7$ & $-15.2$ \\
Kimi K2.6 & External & 32/50 & 420,618 & 130,424 & 290,194 & 69.0\% & $-5.0$ & $+19.8$ & $+5.6$ \\
 & Self & 31/50 & 375,345 & 79,191 & 296,154 & 78.9\% & $-18.9$ & $-4.6$ & $-47.6$ \\
\bottomrule
\end{tabular}
\caption{Retrospective conditional retained-run comparison on all 50 assigned
tasks per arm, including unsolved and model-limit outcomes. Normal scores are
32/50, 30/50, 29/50, and 31/50 in target order. History words count cache-distinct
blocks within each retained run; tool calls use selected terminal trajectories.
Spend includes rewriting. $\dagger$ marks common-tariff target estimates with
actual captured compressor charges. $^{*}$Target time excludes rewriting,
tools, orchestration, and evaluation. Rewrite and inclusive timing appear in
Appendix~\ref{app:generation-speed}.}
\label{tab:deployment-quality}
\end{table*}

\subsection{Coding-Agent Deployment}
\label{sec:deployment}
\paragraph{Setup and accounting.}
We compare continuous compaction with normal history on a fixed 50-task
SWE-bench Verified Mini sample \citep{jimenez2024swebench}. Four targets use
GPT 5.6 Luna as an external compressor or compact their own reasoning. Before
each target request, new reasoning blocks are compacted and cached using only
their text and action name. Blocks of at most 12 words stay unchanged;
non-shorter rewrites revert to the original.

We reanalyze existing captures while keeping the original evaluated run for
every task, including unsolved runs and model-limit exits. The accounting
links saved target responses and associated rewrites to the same run. Earlier
abandoned attempts are excluded. Appendix~\ref{app:generation-speed} defines
this conditional accounting; Appendix~\ref{app:acquisition-overhead} explains
excluded requests and accounting limits.

Spend includes rewriting; the main table reports target time excluding
rewriting. We sum client-observed request
durations, including queueing and network transfer, but exclude tools,
orchestration, and evaluation. These sums do not measure elapsed task time. Nemotron
target usage uses a common tariff, with captured compressor charges. Provider
changes remain part of these comparisons.

\paragraph{Results and practical interpretation.}
Six of eight compaction configurations cost less, and seven reduce target
request time (Table~\ref{tab:deployment-quality}). Only the two MiniMax
configurations reduce rewrite-inclusive request time
(Appendix~\ref{app:generation-speed}). Kimi self-compaction has a small
inclusive time increase that rounds to zero. Rewriting can outweigh time saved on the target's
own requests. A self-compressor may generate substantial reasoning before it
returns a short block. Table~\ref{tab:compressor-effort-main} reports an
exploratory test of lower compressor effort.

Observed success differs from Normal by at most three of 50 tasks. These
results do not establish quality preservation. Paired task-bootstrap intervals
in Appendix~\ref{app:deployment-uncertainty} describe uncertainty over the
assigned tasks while holding the captured runs fixed.

\begin{table}[t]
\centering
\small
\setlength{\tabcolsep}{5pt}
\begin{tabular}{lrr}
\toprule
Compressor effort & Time (s) & Sol Pass / 20 \\
\midrule
Historical xhigh & 90.1 & 12; 13 \\
None & 1.7 & 11 \\
Low & 9.1 & 10 \\
\bottomrule
\end{tabular}
\caption{Twenty selected slow DeepSeek requests.
Historical passes refer to the none/low judging rounds. Median times are
confounded by provider and possible checkpoint differences
(Appendix~\ref{app:compressor-effort}).}
\label{tab:compressor-effort-main}
\end{table}

To understand differences in agent work, we retrospectively count tool calls
in the selected terminal trajectories, excluding abandoned attempts. Changes
range from $-29.8\%$ to $+19.8\%$. Kimi with external compaction illustrates
the combination of more steps and shorter reasoning: selected assistant responses
rise from 1,842 to 2,304 ($+25.1\%$), while mean reasoning per response falls
31.1\% and total reasoning falls 13.8\%. This arithmetic decomposition explains
the total change through response frequency and reasoning length per response.

Preprocessing amortization and single-task marginal latency are analyzed
separately in Appendix~\ref{app:latency}; the hosted one-use example loses both
time and money.

\section{Discussion}
Our controlled comparisons show that changing how earlier reasoning is written
can reduce how much reasoning a model generates next. The target receives no
new instruction to be brief. We still do not know which changes cause the
shortening. Rewrites may remove repetition, emphasize different facts, or lose
important doubts. Changes in later actions can also increase or reduce the
eventual savings.

The human audit highlights the need to explain task
terminology to raters.

For software agents, the goal is to solve tasks reliably with less total work.
Shorter reasoning helps when the savings exceed the cost of rewriting and any
additional work the agent performs afterward. Because these tradeoffs differ
across models, useful compaction policies must be evaluated for task success,
total cost, and time spent on both rewriting and task completion.

Our deployment rewrites new reasoning blocks before each target request. The
compressor sees only the block and its action name. Future work should test
rewriting less often, such as every $N$ agent turns, and giving the compressor
more of the conversation. This additional context may help preserve important
information. When the target also performs compression, shared conversation
prefixes may allow reuse of cached computation. These alternatives should be
compared at similar task quality, counting the cost and time of rewriting.

The exploratory compressor-effort study motivates a
larger matched deployment comparison, keeping the target's task-solving
settings unchanged. Such a study should sample beyond unusually slow requests,
compare compressor effort settings on the same checkpoint and provider, and
measure information retention, task success, and rewrite-inclusive cost and
time.

\begin{samepage}
\section{Conclusion}
Our goal is to express useful reasoning in fewer tokens through a more compact
style. We test whether rewriting earlier scratchpads can encourage this
behavior without retraining. We find frequent shortening among accepted
histories and cost savings in several tested software-agent deployments.
Further work should examine how rewriting affects later token use, when to
rewrite, and what context the compressor needs. More broadly, open-weight
models may already support compact reasoning styles that their default outputs
do not reveal, motivating further work on encouraging these capabilities
through inference-time interventions or lightweight training.
\end{samepage}

\section*{Ethical Considerations}
External rewriting can expose sensitive history to another provider.
Applications should minimize data, redact secrets, protect verbatim fields, and
use trusted or local compressors. Potential loss of uncertainty or verification
state warrants task-specific quality checks.
Human ratings omit participant names.

\section*{Limitations}
The controlled tasks share a narrow structure, source acceptance is asymmetric,
and the same model writes and judges rewrites.
The human audit uses a small accepted subset,
nonexpert family raters, and a post-start extension. Equal-horizon selection limits trajectory-wide inference. We lack a
matched deletion, truncation, or extractive-retention baseline, so comparative
value over free alternatives remains untested. The deployment uses one
nonrandom sample without repeated seeds; mixed providers further limit causal
interpretation. The style and coding studies are exploratory. Visible reasoning
tokens and client-observed request durations leave hidden computation and
complete operational latency outside the measurement scope. Quality
noninferiority remains unestablished. Training-data diversity's effect on
model-family compatibility is untested (Appendix~\ref{app:family-compatibility}).

\bibliography{references}
\clearpage
\appendix

\section{Controlled Protocol and Fidelity Checks}
\label{app:protocol}
\subsection{Task structure and fork delivery}
The deterministic tasks require successive actions and decisions under hard
constraints, updated measurements, and tie-breaking rules. Each source contains
three shared acquisition turns. A reasoning rewrite is generated before the
corresponding acquisition tool executes. Clean and compact histories therefore
share the acquisition actions and observations, with differences restricted to
assistant reasoning. Continuations are sampled only after source acceptance and
history freezing.

Provider-specific historical fields are stripped so the original reasoning
cannot survive beside rewritten plaintext. In controlled forks, both arms use
the same assistant reasoning field, and all other first-request fields match.
Arm order is deterministically shuffled from the frozen continuation seed.
Hosted routes can remain nondeterministic despite identical seed settings.
Action-recovery checks qualify routes by demonstrating functional use of
prior reasoning.

The six routes are DeepSeek V4 Flash/DeepInfra/xhigh, Laguna S
2.1/Poolside/high, MiniMax M3/DeepInfra/high, GLM 4.7 Flash/DeepInfra/high,
GLM 5.1/Z.AI/high, and DeepSeek V4 Pro/Together/xhigh. All use an 8,192-token
completion cap without provider fallback. Main Table~\ref{tab:controlled} uses
a retrospective retry-augmented synthesis. Its replacements comprise three
clean Laguna S 2.1 branches that reached the 8,192-token cap: grid decision
replicate two, manufacturing recall replicate three, and rescue decision
replicate two. A fourth replacement retries the compact rescue replicate two
after a provider-502 response. The rule permits up to 20 attempts per turn,
resending the same captured request and extending a recovered branch thereafter.
The recovery runs use 3, 3, 5, and 3 request attempts, respectively, including
later turns but excluding the original captures.

The three cap events are observed model outcomes under the stated budget; we
have no independent evidence that they were infrastructure faults. Retrying
them selects different continuation outcomes and changes the estimand. The
first-attempt sensitivity below therefore retains all four original branches.
Correct reasoning followed by a nonexact final string still fails the original
strict parser in both summaries.

\subsection{Acceptance and attrition}
\label{app:cohorts}
All routes attempt the same 12 tasks, but retain model-specific histories.
DeepSeek V4 Flash and Laguna retain 12 each, MiniMax M3 retains 10, GLM 4.7
Flash retains five, GLM 5.1 retains 10, and DeepSeek V4 Pro retains nine. Five
source-action failures and nine fidelity exclusions account for the 14 omitted
histories in the final six-route synthesis. Every accepted source contributes
three continuation pairs, including failures and length increases. Two
unequal-turn-count pairs from GLM 4.7 Flash remain in the 174 quality pairs but
are excluded from the 172 comparable token pairs.

These final retention counts do not include every superseded development
artifact. During initial source acquisition, three rewrites failed semantic
review: two changed estimate status to binding-quote status, and one added a
spurious threshold when a detector mistook a phase number for a constraint.
Corrections made after the corresponding tool result had executed were archived
and excluded from continuation sampling, including corrections prompted with
the result withheld.
After fixing the detector and evidence-status wording, the affected sources
were reacquired. A further ambiguous manufacturing source was rejected before a
later reacquisition passed. Seven superseded artifacts are recorded in the
initial exclusion manifest.

\subsection{Rewrite instructions and retries}
The initial acquisition instructions are reproduced in
Appendix~\ref{app:dose} alongside the telegraphic histories reused in the style
sweep. Additional-route prompts use the evidence-status variant there, inserting
the following instructions immediately after its evidence-status sentence:

\begin{quote}\small\raggedright
The requested tool call and its arguments are preserved separately outside the
scratchpad. Treat them only as context: retain the intended next action, but
never import facts, code, or argument details absent from the raw reasoning.
Aim for at most half the raw word count when the state permits, and never make
the scratchpad longer.
Preserve the raw notation rather than translating mathematical expressions into
implementation syntax or otherwise improving them.
\end{quote}

The rewriter receives raw reasoning and the requested tool function and
arguments. It can make up to three draft attempts before that tool executes.
Empty drafts raise the completion cap; missing required candidate identifiers
or hard-threshold values trigger a repair request containing the raw text,
previous draft, and required fields. Failure after three drafts stops that
source. This mechanical retry is distinct from later semantic review. The
additional-route configuration permits no post-capture repair or replacement
source after rejection. Its durable transport permits at most two retries for
retryable returned HTTP errors; ambiguous unreturned requests are not
automatically resent.

\subsection{Review procedure and rubric}
GPT 5.6 Sol produces the controlled rewrites and reviews them in separate
calls. The review payload contains the raw reasoning, candidate state, and
requested tool function and arguments. Future results and continuation outcomes
are excluded from that payload. The rewrite itself precedes tool execution;
source-level review and acceptance precede continuation sampling. These are
distinct timing requirements.

Mechanical checks verify source hashes, reasoning-only differences, requested
actions, candidate identifiers, and hard thresholds. The reviewer checks
material preservation and leakage. A source must have shorter aggregate
reasoning and pass all applicable checks. Identity blocks need no semantic
rewrite review. Model-based acceptance cannot establish independent semantic
equivalence: the generator and judge can share blind spots.
A subsequent human audit examines a small subset of accepted
rewrites (Appendix~\ref{app:human-fidelity}).

The following is the exact semantic-review system prompt used for the
additional-route acquisition. The response schema is part of the prompt.

\begin{quote}\small\raggedright
Audit one compact reasoning-state rewrite before its pending tool result exists.
Pass only if it preserves every decision-relevant fact, number, uncertainty,
conclusion, rejected alternative, evidence-status distinction, and intended next
action from the raw reasoning; adds no unsupported fact; and neither states nor
implies that the requested tool result has occurred. The tool call is separately
preserved and is supplied only to check action consistency. Return only one JSON
object with exactly these keys: verdict ('pass' or 'fail'),
critical\_or\_material\_issue (boolean), future\_result\_leakage (boolean),
requested\_action\_preserved (boolean), explanation (short string).
\end{quote}

Acceptance requires a pass verdict, no material issue, no future-result leakage,
and preservation of the requested action. Malformed or failed review responses
do not pass. The prompt's reference to a nonexistent pending result describes
the state being audited; the reviewer is given no such result. We make no claim
that this model-based procedure replaces an independent annotation protocol.

\paragraph{Treatment example.}
Table~\ref{tab:compaction-example} reproduces a selected
DeepSeek V4 Flash reasoning block and its GPT 5.6 Sol rewrite from turn three
of the vaccine-planning task, before simulation. It illustrates the intended
style change; it is not an independent fidelity assessment.

\begin{table*}[tbp]
\centering
\begingroup
\small
\begin{tabular}{@{}p{0.48\textwidth}@{\hspace{0.04\textwidth}}p{0.48\textwidth}@{}}
\toprule
\textbf{Original reasoning (152 words)} & \textbf{Compact reasoning (41 words)} \\
\midrule
\raggedright
Let me check the hard constraints:
\par\medskip
1. binding\_quote\_usd <= 1,000,000 - all options have estimated\_usd below 1M, but these are estimates, not binding quotes. The hard constraint says "binding\_quote\_usd\_lte" - so we need binding quotes to check this. But we don't have binding quotes yet. Let me proceed.
\par\medskip
2. minimum\_sites >= 2 - V1 has 1 site, so V1 violates this constraint. V2 (3), V3 (2), V4 (3), V5 (2) all satisfy.
\par\medskip
3. reliability\_gte >= 0.94 - we don't have reliability data yet.
\par\medskip
4. risk\_lte <= 0.05 - we don't have risk data yet.
\par\medskip
So we need to simulate to get more data. Let me proceed to simulate. But first, I need to know which options survive static checks. V1 is out due to sites=1 < 2. So the remaining options are V2, V3, V4, V5. But I need exactly 4 option\_ids for simulate\_vaccine\_options. Let me try with V2, V3, V4, V5.
&
\raggedright\arraybackslash
Binding quote $\leq$ \$1,000,000 cannot be verified from estimates. Minimum sites $\geq$2 excludes V1 (1 site); V2 (3), V3 (2), V4 (3), and V5 (2) pass. Reliability $\geq$0.94 and risk $\leq$0.05 lack data. Simulate the four surviving options.
\par\medskip
Requested tool: simulate\_vaccine\_options(\{"case\_id":"VAX-28",\allowbreak"option\_ids":["V2","V3","V4","V5"]\})
\\
\bottomrule
\end{tabular}
\endgroup
\caption{Selected original/rewrite pair from the controlled
experiment. Wording is reproduced verbatim with typeset mathematical symbols
and adjusted line wrapping. Counts are recorded words, not tokens. The rewrite
retains the unresolved quote, reliability, and risk checks, the site-based
rejection, and the intended simulation call. Tool calls and observations are
preserved separately in both histories.}
\label{tab:compaction-example}
\end{table*}

\begingroup
\subsection{Human fidelity audit}
\label{app:human-fidelity}
The raters were two family members of the author without ML or coding
expertise. They compared Original and Rewrite text independently, checking
facts, constraints, uncertainty, and intended actions. Pass means no material
difference found; Problem records an apparent material omission, change, or
unsupported addition; Unsure records an unresolved judgment. Task correctness
and continuation outcomes were outside the assessment.

The initial sampling frame contained 58 accepted source histories. Source and
block selection did not use continuation outcomes, and each rater received a
separately shuffled order. The extension used previously unused histories from
the original route-balanced sample, selecting their third pre-fork assistant
blocks for longer candidate/constraint comparisons. These were presented
longest-first to A and shuffled for B. This extension was selected during data
collection, after A's initial ratings had been received.

In the initial batch, B marked the DeepSeek V4 Flash recall block Unsure:
``I don't know what `pending tool result' means. Next action is omitted.''
The rewrite reads ``Load recall brief for case RECALL-62. Pending tool result.''
B marked the Laguna water-planning block Problem with an omission flag,
quoting ``Now phase 2: list all candidate options.'' Its rewrite specifies
\texttt{list\_water\_options} as the next action. In the extension, B marked the
Laguna flood-planning block Problem with an omission flag, quoting ``phase 3'';
its rewrite specifies \texttt{simulate\_flood\_options} and the surviving option
IDs. These submitted labels remain unchanged. The accompanying text permits
inspection of the reported concerns without introducing consensus labels.

We report exact label agreement and omit Cohen's kappa because A used only one
category. The small accepted subset, the post-start extension, the raters'
relationship to the author, and uncertainty about task terminology limit the
interpretation of this audit. It provides no general fidelity guarantee or
evaluation of downstream task success.
\endgroup

\subsection{Estimation and distributions}
Controlled route and style intervals use 10,000 source-level bootstrap resamples
with replacement. All replicates from a sampled model--task source remain
together; intervals are the 2.5th and 97.5th percentiles. This resampling unit
does not preserve cross-route dependence from shared task definitions.
Replicates measure continuation variability within the same source history.
Route-specific retention and small source counts restrict population
interpretation.

Figure~\ref{fig:raincloud} displays each comparable pair to expose variation
hidden by route medians. Most mass is below zero on every route, but positive
tails remain. GLM 4.7 Flash shortens in all 13 comparable pairs and also has the
lowest retention rate. The shortening estimate is conditional on route-specific
source acceptance.

\subsection{First-attempt and retry-augmented sensitivity}
\label{app:robustness}
\label{app:first-attempt}
To test sensitivity to retrospective retries, an offline audit re-extracts the
original branches of every accepted source and compares them with the
retry-augmented cohort. Here first-attempt means the original captured
continuation before the four retrospective replacements. Acquisition and
transport retries within those captures remain included.

First-attempt captures contain 169 completed equal-horizon pairs, of which 149
shorten, with a paired median reasoning change of $-32.1\%$. Only Laguna's
route-level results change: its original 33 comparable pairs have median
$-47.1\%$ and exact success 33/36$\to$34/36; all other routes retain the values
in Table~\ref{tab:controlled}. Across all 174
pairs, first-turn reasoning changes by $-20.7\%$ at the paired median, and
pooled full-trajectory reasoning falls 38.1\%. Table~\ref{tab:assigned-workload}
retains capped and failed branches. The qualitative shortening survives. Retry augmentation changes failure counts
and workload by replacing three length-capped outcomes and one provider error.

The retry-augmented first requests satisfy the reasoning-only invariant and
share the preceding result in all 174 pairs. First-turn reasoning has a
$-20.0\%$ paired median and a $-16.4\%$ pooled change; subsequent reasoning has
a $-37.8\%$ paired median and a $-40.0\%$ pooled change. Later observations are
also equal in all 173 pairs reaching turn two in both arms and all 172 reaching
turn three. Later generation nevertheless follows treatment-affected text and
actions, so these contrasts can contain mediated effects.

Across the 172 equal-horizon pairs, the median reasoning change is $-32.0\%$
(source-clustered 95\% CI $[-38.8,-24.3]\%$), and the median absolute change
is $-178$ tokens (CI $[-224,-126]$). Resampling the 12 shared task definitions
instead, retaining all accepted routes and replicates within each sampled task,
gives a median-change interval of $[-37.4,-26.0]\%$. This sensitivity preserves
cross-route task dependence and observed gate missingness. The largest increase
is 2,807 tokens. Table~\ref{tab:assigned-workload} retains unequal-horizon pairs and the failures
remaining after retry augmentation. Reasoning falls in aggregate even without
horizon matching. These retry-augmented totals exclude the discarded retry
attempts.

\begin{table*}[tbp]
\centering
\small
\setlength{\tabcolsep}{7pt}
\begin{tabular}{lrrrr}
\toprule
& \multicolumn{2}{c}{First-attempt captures} & \multicolumn{2}{c}{Retry-augmented captures} \\
\cmidrule(lr){2-3}\cmidrule(lr){4-5}
Assigned-cohort quantity & Clean & Compact & Clean & Compact \\
\midrule
Reasoning tokens & 152,475 & 94,357 & 135,778 & 97,790 \\
Accumulated input tokens & 1,399,681 & 1,259,755 & 1,429,111 & 1,277,229 \\
Completion tokens & 234,817 & 159,887 & 220,049 & 163,876 \\
Continuation turns & 515 & 518 & 521 & 520 \\
Exact failures & 12 & 9 & 9 & 8 \\
Action-failure terminations & 4 & 2 & 1 & 1 \\
\bottomrule
\end{tabular}
\caption{All 174 pairs in each cohort definition, without horizon filtering.
First-attempt captures retain cap events and the provider failure; retry-augmented
columns substitute recovered branches and exclude discarded retry work. Input
sums repeated contexts; completion includes reasoning and visible answer/tool
output. Action failures are included among exact failures.}
\label{tab:assigned-workload}
\end{table*}

To assess whether one route drives the result, we omit each route in turn
(Table~\ref{tab:loo}). The median token change stays negative throughout,
between $-36.7\%$ and $-29.2\%$. Omitting GLM 4.7 Flash reverses the quality
change from a net gain to a net loss. Thus shortening is robust to this route
omission check, whereas the aggregate quality sign is not. These source-level
summaries do not estimate new routes, rejected sources, or unobserved provider
and seed variation.

\begin{table*}[tbp]
\centering
\small
\setlength{\tabcolsep}{6pt}
\begin{tabular}{lrrrrr}
\toprule
Omitted route & Shorter/comparable & Median $\Delta$ & Median token $\Delta$ & Success C$\to$K & Quality $\Delta$ (pp) \\
\midrule
DeepSeek V4 Flash & 121/136 & $-32.7\%$ & $-188$ & 129$\to$131 & $+1.4$ \\
Laguna S 2.1 & 119/136 & $-29.2\%$ & $-156$ & 129$\to$131 & $+1.4$ \\
MiniMax M3 & 126/142 & $-33.7\%$ & $-207$ & 135$\to$136 & $+0.7$ \\
GLM 4.7 Flash & 138/159 & $-29.3\%$ & $-160$ & 154$\to$152 & $-1.3$ \\
GLM 5.1 & 123/142 & $-36.7\%$ & $-202$ & 140$\to$141 & $+0.7$ \\
DeepSeek V4 Pro & 128/145 & $-31.8\%$ & $-180$ & 138$\to$139 & $+0.7$ \\
\bottomrule
\end{tabular}
\caption{Retry-augmented leave-one-route-out sensitivity. Token summaries use
equal-horizon pairs; quality uses all assigned pairs on the remaining routes (138, 138, 144,
159, 144, and 147, respectively). C/K denote clean/compact.}
\label{tab:loo}
\end{table*}

\section{Exploratory Controls and Behavioral Case Study}
\label{app:secondary}
\subsection{Length-matched and sentence-style controls}
\begin{figure}[tbp]
\centering
\includegraphics[width=\columnwidth]{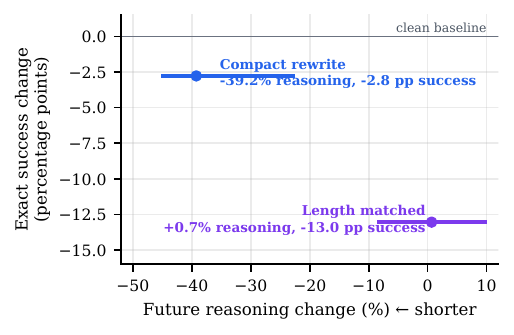}
\caption{Length and quality control. Horizontal position is paired median future
reasoning change, with source-clustered 95\% intervals; vertical position is
exact-success change. The paraphrase control is quality-degrading and therefore
does not isolate history length.}
\label{fig:length-matched-control}
\end{figure}

To test whether rewording alone reproduces shortening, the length-matched
control uses fresh clean and paraphrased continuations from pre-outcome frozen
variants. Paraphrases retain 90--110\% of clean word count. Of 65 comparable
pairs, 31 shorten; the median change is $+0.7\%$, with source-clustered 95\% CI
$[-8.2,+9.7]\%$. Exact success falls from 68/69 to 59/69. For comparison, the
retry-augmented two-route cohort has median $-39.2\%$ across 72 pairs, with
interval $[-44.8,-22.9]\%$ (Figure~\ref{fig:length-matched-control}). These are
separate cohorts, and the control's quality loss prevents attributing the
contrast solely to history length.

To probe syntax within compact histories, another fresh comparison replaces
telegraphic states with full sentences. Of 60 pairs, 33 shorten (55.0\%);
median change is $-4.7\%$, with interval $[-16.9,+9.2]\%$, and exact success
changes from 58/60 to 56/60. Nineteen parent sources are shared by both control
studies. On that common-source subset, paraphrase and sentence-conversion
medians are $-2.4\%$ and $-5.9\%$, respectively, and both intervals include
zero. These controls do not identify a single linguistic cause: syntax and
lexical realization can also alter salience.

\subsection{Selected contest-code continuations}
\label{app:code}
To examine what shorter reasoning can contain, we use a LiveCodeBench v6 screen
with DeepSeek V4 Flash and Laguna S 2.1. Shared acquisition succeeds for 12/24
model--item sources; 12 length-truncated failures are retained in the acquisition
record and are not resampled. The accepted sources yield 36 continuation pairs.
Only 33 have comparable horizons, and these are also the successful pairs. The
screen is therefore conditional on acquisition and a favorable continuation
regime.

Table~\ref{tab:code} gives token and quality outcomes. Across comparable pairs,
29/33 shorten, with median $-57.1\%$. Both arms solve 33/36 continuations under
the official suite. Among successful DeepSeek branches, extended submit-turn
reverification, defined after acquisition as at least 350 reasoning tokens,
occurs in 11/18 clean branches (61.1\%) and none of the compact branches.
Median final-turn reasoning falls from 489 to 12 tokens in the selected screen.

Across successful branches, 32 clean and 33 compact submissions preserve the
shared candidate unchanged. Three unsuccessful DeepSeek pairs repeatedly call
the earlier candidate-recording action after deriving corrected code. These
observations concern verification and termination after shared acquisition;
they cannot establish general repair ability or distinguish harmless checking
removal from lost uncertainty. Passing public tests alone is not evidence that
all further verification is redundant.

\begin{table}[tbp]
\centering
\small
\setlength{\tabcolsep}{3pt}
\begin{tabular}{lrrr}
\toprule
Route & Shorter & Median $\Delta$ & Success C$\to$K \\
\midrule
DeepSeek & 15/18 & $-93.9\%$ & 18/21$\to$18/21 \\
Laguna & 14/15 & $-38.5\%$ & 15/15$\to$15/15 \\
Combined & 29/33 & $-57.1\%$ & 33/36$\to$33/36 \\
\bottomrule
\end{tabular}
\caption{Acquisition-selected coding screen. Source-clustered 95\% intervals
for the medians are $[-97.4,-2.4]\%$ (DeepSeek), $[-58.3,-29.2]\%$ (Laguna),
and $[-95.9,-19.0]\%$ (combined). C/K denote clean/compact.}
\label{tab:code}
\end{table}

\section{Retrospective Style Sweep and Frozen Prompts}
\label{app:dose}
\subsection{Design and interpretation}
To explore whether terser state records are associated with shorter future
reasoning, ten outcome-known DeepSeek source histories each receive three fresh
randomized continuation blocks under five history styles. Relative to clean,
median future reasoning changes are $-24.5\%$ for relaxed prose, $-27.0\%$ for
compact full sentences, $-30.1\%$ for telegraphic states, and $-37.4\%$ for
ultra-telegraphic states. Source-clustered 95\% intervals are
$[-37.7,-12.5]\%$, $[-31.6,-17.6]\%$, $[-41.0,-23.8]\%$, and
$[-41.9,-23.0]\%$. Every arm has 100\% exact success.

Only 6/30 continuation blocks are monotonic across all tiers. Adjacent median
contrasts are $-4.9\%$, $-6.8\%$, and $-9.6\%$, and two of their three
intervals include zero. Across 30 continuation blocks from ten source histories, the median Spearman
correlation between history word count and future reasoning tokens is 0.70. Ordered arm medians therefore support an
exploratory association at the group level. Length, syntax, wording, and
compressor provenance vary jointly.

Figure~\ref{fig:controlled-summary}(b) in the main text
summarizes this style sweep; panel (a) reports a separate turnwise analysis of
the initial two-route comparison.

\subsection{Frozen rewrite prompts}
Several simple prompt formulations were tried during development. A model
generated the frozen wording before evaluation; variants were not selected using
measured compression, later reasoning, or task outcomes. The relaxed-prose,
compact-full-sentence, and ultra-telegraphic treatments used the following common
system prompt, paired with the exact style directive and turn-specific source
fields.

\begin{quote}\small\raggedright
Rewrite one historical assistant reasoning block into the requested style tier.
Output only the rewritten state through the required JSON schema.
Preserve every decision-relevant fact, number, unit, comparator, uncertainty,
correction, rejected branch, evidence-status distinction, conclusion, dependency,
and intended next action supported by the raw scratchpad and accepted compact
anchor. The raw scratchpad is authoritative if the anchor adds process-control
metadata; do not restore irrelevant repetition.
Include the requested action's exact function name, argument keys, and argument
values in the state. The action has not executed: do not add process metadata
saying its result is pending, withheld, absent, or forthcoming.
Preserve an unknown/pending measurement only when the reasoning texts themselves
state it. State every inequality direction and tie-break direction explicitly;
`limit X to N' or a bare metric name is not a substitute for $\leq$/$\geq$,
lower/higher, at-most/at-least, minimum/maximum, or the source's exact direction.
Never import a fact absent from both reasoning texts, improve the solution, use
a later turn, observation, or final answer. Preserve opaque identifiers exactly;
do not invent shorthand such as P2 for Phase 2 or DAY-21 for day 21.
Stay inside the exact inclusive word interval. Do not emit template-control
tags, code fences, labels, commentary, or an unexplained token list.
\end{quote}

\paragraph{Relaxed or casual prose.}
\begin{quote}\small\raggedright
Express the state as relaxed, natural, complete prose with modest connective
context. Remain materially shorter than the raw scratchpad. Never repeat facts
merely to reach the interval.
\end{quote}

\paragraph{Compact full sentences.}
\begin{quote}\small\raggedright
Express the compact state in ordinary complete grammatical sentences. Make
dependencies explicit without adding explanation, repetition, or facts. Avoid
telegraphic fragments and note headings.
\end{quote}

\paragraph{Ultra-telegraphic states.}
\begin{quote}\small\raggedright
Write the shortest unambiguous state record possible. Use fragments, symbols,
semicolons, and shared modifiers aggressively. Remove rhetoric, repeated rules,
repeated arithmetic, headings, and prose that does not change the next decision.
Omit no decision-relevant state.
\end{quote}

The telegraphic treatment reused accepted parent rewrites. Nine source histories used this frozen system prompt:

\begin{quote}\small\raggedright
Rewrite an assistant reasoning scratchpad into a shorter state record.
Output only the rewritten scratchpad, with no preamble or code fence.
Preserve every decision-relevant fact, number, uncertainty, conclusion, and
requested next action.
Do not add knowledge, improve the solution, or predict the requested tool's
result: that tool has not run yet.
Remove restatement, headings, rhetorical transitions, and repeated arithmetic.
Fragments, equations, and compact clauses are allowed.
\end{quote}

One reacquired source history used this variant:

\begin{quote}\small\raggedright
Rewrite an assistant reasoning scratchpad into a shorter state record.
Output only the rewritten scratchpad, with no preamble or code fence.
Preserve every decision-relevant fact, number, uncertainty, conclusion, and
requested next action.
Preserve evidence-status distinctions exactly, including estimate versus binding
quote and observed versus pending.
Do not add knowledge, improve the solution, or predict the requested tool's
result: that tool has not run yet.
Remove restatement, headings, rhetorical transitions, and repeated arithmetic.
Fragments, equations, and compact clauses are allowed.
\end{quote}

\section{Continuous-Compaction Deployment Details}
\label{app:swe}
\subsection{Assigned cohort and history policy}
The fixed Verified Mini sample contains 25 Django and 25 Sphinx tasks, selected
by an upstream size-optimized LP/clustering procedure. The dataset revision is
\texttt{b316c349947c29963fce3}\allowbreak\texttt{f4a65967c9807a4b673}.
We use upstream mini-SWE-agent 2.4.5 with a bash tool and 250-step contract,
and the official SWE-bench 4.1.0 evaluator.

Before each target request, every unique prior nonempty reasoning block is
compacted once and cached by its raw text and requested action name. Tool
arguments and results remain outside the compressor payload. GPT 5.6
Luna/OpenAI at high effort is the external compressor; self-compaction uses the
target model and route. Blocks of at most 12 words stay identical. A returned
draft that is not shorter reverts to the raw block. Identifier and numeric
omissions are recorded as lexical diagnostics. All tasks remain eligible;
semantic failure requires separate assessment.

DeepSeek and MiniMax replay native reasoning history. Kimi and Nemotron use
symmetric tagged-visible history in both arms because their tested transport
lacks the same native replay contract. Those comparisons concern the symmetric
visible representation.

The first model-terminal prediction is retained per task. Terminal statuses
include submission, limit exit, repeated format error, and timeout.
Acquisition can retry infrastructure/provider failures; completed calls in
abandoned attempts are excluded from the retained-run totals. Provider transitions retain already selected
predictions. Nemotron acquisition moves from NVIDIA's free endpoint to
DeepInfra and BaseTen; Kimi uses Baidu, CoreWeave, and Chutes. These transitions
leave serving variation in the retained runs and preclude a
provider-controlled causal interpretation.

\subsection{Compression and selected tool calls}
\label{app:agent-steps}
Main Table~\ref{tab:deployment-quality} uses the same retained runs as the primary
cost and time comparison. Word counters come from serialized compaction events
in the hash-matched retained trajectories. Across eight treatment arms,
compaction removes 1,072,439 of 1,676,481 history-block words (64.0\%). Each
raw-block/action pair counts once within its retained run. Identity rewrites
remain included, and the same block can count again in another run.

These counts exclude abandoned attempts and omit repeated prompt replay and
non-reasoning context. They cannot substitute for billed input tokens. Tool
calls count assistant-issued calls in the terminal trajectory selected for
evaluation, excluding compressor calls and abandoned attempts.

Tool-call analysis was defined retrospectively; neither target nor compressor
was instructed to reduce tool use. In the four DeepSeek and MiniMax treatment
arms, selected response counts fall 22.6--30.9\%, mean reasoning per response
falls 17.0--30.9\%, and total reasoning falls 36.6--49.2\%. All 50 selected
DeepSeek trajectories submit. MiniMax submission counts are 38 under normal
history, 42 under external compaction, and 44 under self-compaction, with
correspondingly fewer limit or repeated-format exits. These describe changed
iteration and termination. Their causal mechanisms remain unresolved.

\paragraph{Kimi's increased tool use.}
\label{app:kimi-tool-calls}
To check the positive tool-call exception, we inspect all selected Kimi
external-compaction trajectories. Calls rise from 1,990 to 2,384, a net increase
of 394 (19.8\%). They rise on 29 tasks, tie on three, and fall on 18; the
paired median change is $+9.3\%$. The three largest increases contribute 208
calls. Selected reasoning falls from 895,162 to 771,407 tokens ($-13.8\%$),
while assistant responses rise from 1,842 to 2,304 ($+25.1\%$) and mean
reasoning per response falls 31.1\%. More responses with shorter reasoning
account arithmetically for the aggregate pattern. No internal analysis or
style-matching experiment identifies its cause.

\subsection{Model-family compatibility hypothesis}
\label{app:family-compatibility}
One possible explanation for target-specific compaction effects is familiarity
with the reasoning styles encountered during training. If a model's reasoning
training relies mainly on self-generated traces or distillation from teachers
in the same family, it may encounter limited diversity in how reasoning states
are expressed. Under this hypothesis, self-compaction could preserve familiar
conventions, while a different family's rewrites could change the cues used to
resume reasoning and decide whether further checking is needed.

The Kimi deployment motivates this question: self-compaction incurs less total
spending and fewer selected tool calls than GPT 5.6 Luna compaction
(Table~\ref{tab:deployment-quality}). The external arm resolves slightly more
tasks in this sample, so this pattern concerns operational efficiency. We do
not measure training-data diversity or teacher provenance. Rewrite length,
information retention, compressor capability and reasoning effort, caching,
and provider variation also remain possible explanations.

Testing style compatibility requires crossing target models and compressors
from the same and different families, with matched rewrite lengths,
independently assessed state retention, and controlled inference settings.
Content-matched stylistic variants could help separate familiarity from
information loss. Testing the training-data explanation itself would require
known teacher provenance or models trained on controlled mixtures of
single-family and diverse-family reasoning traces. The present results do not
establish this mechanism or any target's training composition.

\subsection{Paired task outcomes and uncertainty}
\label{app:deployment-uncertainty}
To quantify uncertainty without treating repeated configurations as independent
tasks, an offline audit resamples the 50 task IDs as pairs within each
model--compressor comparison. Table~\ref{tab:discordant} gives 10,000-draw
paired-bootstrap intervals and discordant resolved sets. A gain is resolved
only under compaction; a loss is resolved only under normal history. Intervals
are conditional on the fixed selected sample and one captured run per arm,
omitting unobserved seed and provider variation. These few discordant tasks do
not establish preservation or improvement. Percentile intervals for arms with
only gains or only losses have one-sided support and cannot sample an absent
direction of discordance. Repeated appearances of tasks across configurations are not pooled as
independent observations.

\begin{table*}[tbp]
\centering
\small
\setlength{\tabcolsep}{7pt}
\begin{tabular}{llrrr}
\toprule
Target & Compressor & Gain/loss & Accuracy $\Delta$ (pp) & Paired-bootstrap 95\% CI (pp) \\
\midrule
DeepSeek V4 Flash & External & 1/1 & $0$ & $[-6,6]$ \\
 & Self & 0/3 & $-6$ & $[-14,0]$ \\
MiniMax M3 & External & 3/2 & $+2$ & $[-6,10]$ \\
 & Self & 3/3 & $0$ & $[-10,10]$ \\
Nemotron 3 Ultra & External & 3/0 & $+6$ & $[0,14]$ \\
 & Self & 2/2 & $0$ & $[-8,8]$ \\
Kimi K2.6 & External & 4/3 & $+2$ & $[-8,12]$ \\
 & Self & 3/3 & $0$ & $[-10,10]$ \\
\bottomrule
\end{tabular}
\caption{Paired resolution differences on 50 assigned tasks per configuration.
The intervals describe task resampling with each captured run held fixed.}
\label{tab:discordant}
\end{table*}

\subsection{Retained-run linkage, cost, and request time}
\label{app:generation-speed}
This retrospective reanalysis uses existing captures without rerunning agents
or evaluations. The retained attempt follows the original first model-terminal
outcome rule with a saved trajectory: Submitted, LimitsExceeded,
RepeatedFormatError, or TimeExceeded. All 50 assigned tasks per arm remain,
regardless of official success. All 400 treatment patch hashes match the frozen
evaluation prediction bindings. Complete retained target response-ID hash sets
match the frozen selected-response ledger for every treatment arm. These are
provenance checks on existing outcomes, not new experimental trials.

Target cost and duration use completed captured responses matched by those
IDs. Rewrite responses are taken from the compressor subdirectory of each
retained trajectory's \texttt{info.config.durable\_journal\_dir}. Attempt
numbers alone cannot bind calls to tasks because they repeat across tasks.
Per-trajectory rewrite charges match serialized compaction cost totals.
Responses are deduplicated by response ID, or response-body hash if no ID
exists, with rewrite uniqueness checked per arm. Only completed JSON responses
with nonempty \texttt{choices} contribute, including length-limited responses.

Spend adds target and corresponding rewrite charges. Non-Nemotron targets and
all compressors use captured reported charges; Nemotron targets use the common
tariff in Appendix~\ref{app:mixed-provider-spend}. Inclusive request time is the
sum of target and rewrite durations. Client-observed durations include queueing,
prefill, generation, and network transfer. Tools, orchestration, and official
evaluation are excluded. Concurrency separates summed request-hours from elapsed
wall time; no complete rewrite-inclusive task wall-time comparison is available.

\begin{table*}[tbp]
\centering
\small
\setlength{\tabcolsep}{3.5pt}
\begin{tabular}{llrrrrr}
\toprule
Target & Arm & Spend (\$) & Target (h) & Rewrite (h) & Inclusive (h) & Inclusive $\Delta$ (\%) \\
\midrule
DeepSeek V4 Flash & Normal & 2.43 & 3.37 & 0.00 & 3.37 & --- \\
 & External & 1.59 & 2.22 & 1.23 & 3.45 & $+2.4$ \\
 & Self & 1.30 & 1.91 & 3.94 & 5.84 & $+73.3$ \\
MiniMax M3 & Normal & 17.72 & 23.90 & 0.00 & 23.90 & --- \\
 & External & 11.70 & 8.86 & 0.90 & 9.76 & $-59.1$ \\
 & Self & 10.95 & 9.38 & 1.44 & 10.82 & $-54.7$ \\
Nemotron 3 Ultra & Normal & 17.67 & 5.89 & 0.00 & 5.89 & --- \\
 & External & 19.33 & 4.27 & 1.84 & 6.10 & $+3.7$ \\
 & Self & 18.95 & 4.99 & 3.52 & 8.51 & $+44.5$ \\
Kimi K2.6 & Normal & 26.77 & 8.96 & 0.00 & 8.96 & --- \\
 & External & 25.43 & 9.47 & 2.06 & 11.52 & $+28.5$ \\
 & Self & 21.73 & 4.70 & 4.27 & 8.97 & $+0.0$ \\
\bottomrule
\end{tabular}
\caption{Retained-run spend and completed request-duration sums over all 50
assigned tasks per arm. Spend and inclusive time include rewriting; inclusive
changes are relative to Normal. Nemotron target charges are common-tariff
estimates with actual captured compressor charges. Tools, orchestration, and
evaluation are excluded. Columns are rounded independently.}
\label{tab:request-hours}
\end{table*}

\subsection{Excluded requests and accounting limits}
\label{app:acquisition-overhead}
The retained-run comparison excludes earlier abandoned attempts and unlinked
work, including completed generations absent from retained trajectories. These
exclusions are not exclusively independently classified provider failures.
They leave every assigned task and its original official outcome in place.
Conditioning on retained runs does not control provider or seed variation and
cannot identify a causal effect of failures or a controlled change in serving
speed.

Returned JSON bodies without nonempty \texttt{choices} are excluded from cost
and request-time totals and have zero captured charges. Uncaptured or unreturned
requests are also excluded; their billing and durations remain unknown. The
retained totals therefore describe captured work linked to selected runs, not
complete acquisition spending or elapsed task time.

\paragraph{Common-tariff Nemotron estimates.}
\label{app:mixed-provider-spend}
To avoid comparing one arm's free quota with another's paid acquisition,
retained completed Nemotron target calls are repriced at common DeepInfra tariffs:
\$0.50 per million uncached input tokens, \$0.10 per million cache-read tokens,
and \$2.20 per million output tokens. Recorded cached tokens receive the
cache-read rate; all recorded cache-write counts are zero. Actual compressor
charges remain included. These estimates use recorded input, cache, and output
usage; common pricing does not remove provider or seed confounding.

\begingroup
\subsection{Exploratory compressor-effort comparison}
\label{app:compressor-effort}
To examine expensive self-compaction calls, we select the 20 slowest completed
DeepSeek V4 Flash compressor responses among 1,567 recorded responses in the
self-compaction deployment ledger. These requests contain 20 distinct original
reasoning blocks. On September 15, 2026, we replay their unchanged system and
user messages with \texttt{deepseek/deepseek-v4-flash-0731} on DeepInfra, using
reasoning effort \texttt{none} and \texttt{low}. The historical \texttt{xhigh}
responses came from Baidu under an unversioned model identifier.

GPT 5.6 Sol, with medium reasoning effort, assesses each new rewrite alongside
its historical counterpart against the original block. The judge sees no arm
identities, timing, or token counts. A/B positions are randomized and held fixed
across the two rounds. Each pair receives one judgment using the same
Pass/Problem/Unsure fidelity rubric. Non-shorter candidates remain in the
analysis. This is a model-based assessment, separate from the human audit.

\begin{table}[tbp]
\centering
\small
\setlength{\tabcolsep}{3pt}
\begin{tabular}{lrrrr}
\toprule
Effort & Time (s) & Reasoning & Shorter & Sol Pass \\
\midrule
xhigh & 90.08 & 10,536 & 18 & 12; 13 \\
none & 1.69 & 0 & 18 & 11 \\
low & 9.06 & 275.5 & 16 & 10 \\
\bottomrule
\end{tabular}
\caption{Exploratory compressor-effort results on
20 selected requests. Time and reasoning tokens are per-request medians.
Shorter counts outputs with fewer whitespace-delimited words than the original;
Sol Pass counts fidelity passes. The historical row gives passes in the
none-comparison and low-comparison rounds, respectively.}
\label{tab:compressor-effort}
\end{table}

All completed outputs end with a stop finish reason. The initial reasoning-off
run used a 4,096-token cap, which no completed response reached, and a
180-second inactivity timeout. One interrupted stream was retained and retried
without those limits; its 182.7-second failed attempt is excluded from the
completed-response timings in Table~\ref{tab:compressor-effort}. The low arm
used no client timeout or output cap. All reasoning-off responses report zero
reasoning tokens.

Low effort does not improve the pass count over reasoning off in this sample.
Four cases change from Problem to Pass and five from Pass to Problem. One
unchanged historical rewrite also receives a different label across judging
rounds, showing variation in the model-based assessment. The remaining
judgments are Problem; no Unsure labels occur. No agent continuations are run.

The historical timing comparison is confounded by provider, acquisition date,
and potentially checkpoint version. There is no contemporaneous xhigh arm, and
selection of unusually slow requests prevents estimation of typical deployment
savings. These results motivate a larger matched study with broader source
sampling and downstream task evaluation.
\endgroup

\section{Preprocessing and Marginal Latency}
\label{app:latency}
\subsection{Hosted one-use and reuse accounting}
To determine whether reduced continuation time recovers preprocessing overhead,
a separate analysis uses the original initial-study captures: 69 comparable
pairs from 24 sources, before the retrospective retries used in main
Table~\ref{tab:controlled}. This equal-horizon two-route accounting example is
separate from the full six-route workload comparison.

Continuation-only wall time falls 25.9\% at the paired median (95\%
source-clustered CI $[-38.4,-16.4]\%$), an absolute paired median difference of
$-2.825$ seconds. Adding one source rewrite serially yields a 27.0\% increase
and a $+4.079$-second paired median difference. Allocating its duration equally
across three continuations gives a 5.6\% decrease and a $-0.657$-second
paired median difference. These absolute and relative medians are computed
separately and should not be combined to reconstruct a common baseline median.

Rewriting costs \$0.339, compared with \$0.096 for all initial controlled
continuations. Every one-use case is more expensive; amortization over three
replicates also fails to save money. The median added cost is \$0.012906 per
one-use case and \$0.004223 under three-way allocation. These expensive
controlled rewrites serve the intervention study; the deployment uses a
different external compressor.

For a reusable history, a simple fixed-work accounting model would require
$n(T_C-T_K)>T_R$ to recover rewrite time, and separately
$n(C_C-C_K)>C_R$ to recover rewrite cost. Here $T_R,C_R$ are one-time
rewriting overheads and $T_C,T_K,C_C,C_K$ are matched clean/compact
continuation quantities. A finite break-even requires a positive per-use
saving and assumes it persists under reuse. The reported medians and aggregate
charges do not establish a universal break-even count, especially for money.
The three-replicate allocation models experimental reuse.

A separate source-matched reanalysis uses all three continuations from each of
24 histories to estimate the mean marginal saving per source. Time savings are
positive for 22/24 sources. Among those, the continuous time break-even threshold
has median 1.76 continuations and range 0.43--8.33; 18/24 sources recover
rewriting time by three uses. The other two have no finite time break-even under
their observed means. Monetary marginal savings are positive for all 24, but
the corresponding threshold has median 82.4 continuations and range
31.2--82,725; none recovers rewriting cost by three uses. These are
source-conditional projections under fixed mean savings.

\subsection{Single-task self-hosted marginal timing}
To test whether reduced generation can translate into lower latency on a fixed
serving stack, DeepSeek V4 Flash NVFP4 continues the SWE-bench task
\texttt{sphinx-doc\_\_sphinx-10466} from already available clean, external, and
self-compact histories on one B300 with TP1 and vLLM 0.25.0.
Five dedicated and five shared three-arm blocks constitute the original test.
A separate stress test attempts 25 three-arm blocks with 64 users and at most
32 active sequences. The primary timing comparison uses the 23 complete original
blocks. Normal block two and self-compaction block 17 fail protocol checks;
their original failures and durations remain in the record. A sensitivity
analysis replaces those two individual arms with later recovery runs. Its 25
rows combine original arms with two noncontemporaneous replacements.

\begin{table}[tbp]
\centering
\small
\setlength{\tabcolsep}{4pt}
\begin{tabular}{lrrr}
\toprule
Regime (background) & Clean (s) & External $\Delta$ & Self $\Delta$ \\
\midrule
Dedicated (0) & 45.821 & $-26.8\%$ & $-25.4\%$ \\
Shared (20) & 73.464 & $-8.6\%$ & $-25.4\%$ \\
Overloaded (61) & 154.469 & $-26.1\%$ & $-23.4\%$ \\
\bottomrule
\end{tabular}
\caption{Marginal postfork task time after compact history is available: five
dedicated, five shared, and 23 complete original overloaded blocks. Clean is the
regime median; treatment entries are paired median changes. Three experimental
clients run in addition to the background clients.}
\label{tab:walltime}
\end{table}

Across ten original blocks, self-compaction shortens every continuation's time
(paired median $-25.4\%$), and external compaction shortens 90\% (median
$-18.5\%$). Among the 23 complete original stress blocks, median changes are
$-23.4\%$ for self and $-26.1\%$ for external compaction. The 25-row
retry-augmented sensitivity gives $-23.4\%$ and $-26.4\%$, respectively.
Completed original branches have recorded resolved evaluations. The normal
recovery lacks its evaluator-report file and enters only the timing sensitivity.
Table~\ref{tab:walltime} reports the original complete-block comparisons.

Postfork time covers model requests, orchestration, tools, patch capture, and
local validation. It excludes rewriting, setup, replay, and hidden evaluation.
The result estimates marginal continuation latency conditional on a prepared
history. Completion here means protocol completion; the timing comparison does
not establish quality preservation. Because continuations perform different
work, the regimes do
not identify a load-response curve. Resampling these blocks measures variation
among repeated continuations of the same task.

\begingroup
\section{Motivating observations}
\label{app:motivating-observations}
Theo's video \emph{Why is OpenAI so much more efficient?}
(June 30, 2026, 19:40--24:04) illustrates the informal observations motivating
this work.\footnote{\url{https://youtu.be/ypO0q_8zhWw?t=1180}}
The segment discusses abbreviated fragments presented as leaked reasoning
traces and suggests that internal notes can use a more compact writing style
than final answers. This could leave more of a fixed token budget for useful
analysis. The provenance of the displayed traces and the proposed explanations
of OpenAI's training remain unverified.
\endgroup

\end{document}